# Large Language Models' Accuracy in Emulating Human Experts' Evaluation of Public Sentiments about Heated Tobacco Products on Social Media: Evaluation Study

Kwanho Kim, PhD[1] & Soojong Kim, PhD[2]

[1] Department of Media, College of Politics and Economics, Kyung Hee University, Republic of Korea

[2] Department of Communication, University of California Davis, United States

Author Note

Soojong Kim  https://orcid.org/0000-0002-1334-5310

Correspondence concerning this article should be addressed to Soojong Kim, 1 Shields Ave, Kerr Hall #361, Davis, CA 95616, United States. Email: sjokim@ucdavis.edu

# Abstract

**Background:** Sentiment analysis of alternative tobacco products discussed on social media is crucial in tobacco control research. Large Language Models (LLMs) are artificial intelligence models that were trained on extensive text data to emulate linguistic patterns of humans. LLMs may hold potentials to streamline the time-consuming and labor-intensive process of human sentiment analysis.

**Objective:** The accuracy of LLMs in replicating human sentiment evaluation of social media messages relevant to heated tobacco products (HTPs) was examined.

**Methods:** GPT-3.5 and GPT-4 Turbo were used to classify 500 Facebook and 500 Twitter messages. Each set consisted of 200 human-labeled anti-HTPs, 200 pro-HTPs, and 100 neutral messages. The models evaluated each message up to 20 times to generate multiple response instances reporting its classification decisions. The majority label from these responses was assigned as a model's decision for the message. The models' classification decisions were then compared to those of human evaluators.

**Results:** GPT-3.5 accurately replicated human sentiment evaluation in 61.2% of Facebook messages and 57.0% of Twitter messages. GPT-4 Turbo demonstrated higher accuracies overall, with 81.7% for Facebook messages and 77.0% for Twitter messages. GPT-4 Turbo's accuracy with three response instances reached 99% of the accuracy achieved with twenty response instances. GPT-4 Turbo's accuracy was higher for human-labeled anti- and pro-HTPs messages compared to neutral messages. Most of GPT-3.5 misclassifications occurred when anti- or pro-HTPs messages were incorrectly classified as neutral or irrelevant by the model, whereas GPT-4 Turbo showed improvements across all sentiment categories and reduced misclassifications, especially in incorrectly categorized messages as irrelevant.

**Conclusions:** LLMs can be utilized in analyzing sentiment in social media messages about HTPs. Results from GPT-4 Turbo suggest that accuracy can reach approximately 80% compared to the results of human experts even with a small number of labeling decisions generated by the model. A potential risk of using LLMs is the misrepresentation of the overall sentiment due to the differences in accuracy across sentiment categories. Although this issue could be reduced with the newer language model, future efforts should explore the mechanisms underlying the discrepancies and how to address them systematically.

# Introduction

Heated tobacco products (HTPs) are emerging tobacco products that heat processed tobacco leaves, enabling users to breathe nicotine into their lungs [1]. As these products gain global market share at a rapid pace, their potential impacts on tobacco prevention and cessation initiatives are becoming an important topic of public debate [2].

Social media platforms are where a wide range of stakeholders of tobacco regulations distribute their messages, such as policy announcements, product ads, and product user feedback [3–5]. Analyses of social media discourses on HTPs provide opportunities to observe and identify the dynamics of these messages, which could affect the public's perception of these products and relevant regulatory issues [3–7].

Sentiment analysis is a widely adopted method to understand the attitudes of the public toward tobacco-related issues by evaluating social media messages [8–11]. Prior research has specifically focused on positive and negative sentiments due to their possible associations with tobacco cessation and prevention outcomes, such as the use of tobacco products and support for tobacco regulations [11–13].

In past sentiment analyses of large-scale social media content data, human evaluators often examined a subset of the dataset rather than analyzing the entire dataset. The subset was then used as a representative sample to inform the sentiment of the whole dataset [8,9] or as reference for machine-learning classifiers tasked with analyzing the entire dataset [14]. This approach stems from the time-consuming and labor-intensive nature of human sentiment evaluation, which involves recruiting, training, and coordinating multiple evaluators. This complexity arises from the fact that latent coding, including sentiment analysis, requires understanding the underlying meanings and subtleties in the text, which can substantially vary depending on the context and across coders [15].

Large Language Models (LLMs), such as OpenAI's Generative Pre-trained Transformer (GPT) and Google's Gemini, may be able to alleviate the burdens of human sentiment evaluation. LLMs are artificial intelligence (AI) models that were trained on extensive text data to emulate the linguistic patterns of humans [16,17]. Recent LLMs are known to achieve precision at the level of human decisions on several intellectual tasks [18]. As LLMs become increasingly accessible and available, expectation is growing about the feasibility of using these technologies in public health and social science research [19]. Several examples include analyzing health and medical information [20,21], pre-testing the effect of health campaign messages [22], predicting psychological experimental results [23], and simulating sociodemographic groups and their reactions to social issues [24].

We investigate the accuracy of LLMs in analyzing sentiment in social media messages about HTPs. The current research focuses on OpenAI's GPT, given its high accessibility, availability, and popularity. GPT models are easily accessible through chatbot services such as ChatGPT, Microsoft Copilot, and Apple Intelligence, and are estimated to have the largest userbase worldwide. For instance, ChatGPT has more than 200 million weekly active users as of August 2024 [25]. These aspects contribute to the attractiveness of GPT as an analytic tool for tobacco researchers, especially those with limited budgets and resources.

The current research examines the accuracy of GPT-3.5 and GPT-4 Turbo in emulating human sentiment evaluations of social media messages related to HTPs. GPT-3.5 is a milestone model that powered ChatGPT when the service was launched [26]; GPT-4 Turbo is one of the most recent GPT-4 models as of 2024, with a particular development focus on improvements in processing text prompts [27]. The present study conducted direct comparisons of the sentiment evaluations made by human coders and these language models, based on social media messages gathered from multiple platforms. This investigation could ultimately contribute to assessing the ability of LLMs in examining how the public views alternative tobacco products.

## Methods

### Data Collection

Messages relevant to HTPs were collected from two social media platforms that provide distinct message formats: Facebook (long format) and Twitter (short format). Facebook posts were collected utilizing CrowdTangle (CT)'s keyword search feature. CT was a social media analytic tool provided by Facebook's parent company, Meta. It allowed researchers to access the historical data of Facebook [28]. Tweets were gathered using Twitter's API 2.0, which could access the historical Twitter dataset via the company's academic research access program [29]. The current research focused on messages written in English.

In April 2022, a keyword search was conducted using the following search query: "heat not burn" OR "heat-not-burn" OR "heated tobacco" OR "tobacco heating" OR ((htp OR hnb) AND (smoking OR smoke OR vaping OR vape OR tobacco OR cig OR nicotine)) OR iqos. This query was design to find tweets and Facebook posts meeting at least one of the following conditions: (1) Containing "heat not burn" in its entirety. (2) Containing "heat-not-burn" in its entirety. (3) Containing "heated tobacco" in its entirety. (4) Containing "tobacco heating" in its entirety. (5) Containing at least one of "htp" and "hnb", only when it also contains one of "smoking", "smoke", "vaping", "vape", "tobacco", "cig", and "nicotine" (6) Containing "iqos". This search yielded 16,284 Facebook posts that were published between January 2014 and December 2021 and 60,031 tweets published in the same period.

## Human Evaluation

The procedures for preparing samples for human sentiment evaluations were adapted from sentiment analyses of tobacco-related mass and social media discussions [8–11]. A team of three human coders evaluated the sentiment of 1,250 Facebook posts and 1,200 tweets sampled from the entire pool of keyword-searched social media messages (i.e., 16,284 Facebook posts and 60,031 tweets). Those messages were human-labeled as one of the following five categories: ANTI (anti-HTPs messages), PRO (pro-HTPs messages), NEU (neutral messages), MIX (messages containing a mixture of positive and negative attitudes on HTPs), and IR (messages not relevant to HTPs).

The messages were sampled through a multi-step process designed to increase the likelihood of including both potentially negative and positive messages, ensuring their inclusion in the selected messages for the human annotation. The details of the sampling and coding procedures are reported in the online supplement.

## GPT Evaluation

From each of the human-evaluated 1,250 long-form and 1,200 short-form messages, we randomly selected 200 PRO, 200 ANTI, and 100 NEU messages, totaling 1,000 messages. All these selected messages for GPT-3.5 and GPT-4 Turbo sentiment classification were relevant to HTPs. A LLM prompt was created for each message, including coding instructions, the message, and the coding scheme. Because the current study aims to conduct direct comparisons between the sentiment evaluations of human coders and the language models, the instructions and the coding scheme for the language models were kept consistent with those for human coders, aside from minor formatting adjustments. The instructions included in the prompt directed a language model to categorize a given message based on the coding scheme and to format its response based on formatting rules. The coding scheme, largely identical to the one given to human evaluators, included the definitions and explanations of HTPs and five sentiment categories (ANTI, PRO, NEU, MIX, and IR).

LLMs generate a sequence of words by selecting each words based on its preceding words, and the selection is done by sampling a word from a large distribution of possible words [16,17]. Because of the inherent randomness in this sampling process, LLMs may produce different responses to the same prompt. This potential variability can be accounted for by generating multiple responses from an LLM using the same prompt [30]. To be specific, we collected 20 instances of responses for each message, referred to as "response instances." Each instance was obtained by initiating a new chat with a language model, sending the prompt, and saving the response from the model.

A language model' decision for each message was determined by randomly selecting $m$ instances from a pool of 20 response instances, with replacement. Then, the majority within the selected instances was identified. This majority outcome was termed the "machine decision." We assessed the machine decision varying the number of response instances ($m$ = 1, 3, 5, 7, 9, and 11). For example, $m$ = 5 simulates a scenario where a user generates five response instances and identifies the majority among them. $m$ = 1 corresponds to a "one-shot" determination, where a single instance was generated and considered as the machine decision. In case of a tie, an extra response instance was randomly selected until the tie was broken.

For each message and each value of $m$, the process of determining a machine decision was iterated 1,000 times. After each iteration, a variable that we refer to as "human-machine concurrence" was recorded as 1 if the machine decision aligned with the human evaluation of the message. Otherwise, it was recorded as 0. This variable was then averaged across all iterations, yielding a value referred to as "accuracy." Thus, the accuracy in the current study indicates how accurately the language models classify the sentiment of a message based on their $m$ number of responses. Alternatively, the accuracy can be interpreted as the proportion of messages classified by the language models that match the human sentiment evaluation of the same messages. For instance, if the accuracy of a language model is 90% for ANTI message classification, this suggests that nine-tenths of human-labeled ANTI messages are categorized as ANTI by the model. To assess the overall accuracy of the model in evaluating sentiment of a specific set of messages (e.g., Facebook messages classified as ANTI by human evaluators), we calculated the average accuracy across messages in each set, denoted as $K_m$. For example, $K_{11}$ for ANTI messages refers to the proportion of human-labeled ANTI messages that were also classified as ANTI by a language model, based on 1,000 iterations of the majority of randomly selected 11 responses out of the total of 20 responses. Examples of human-labeled ANTI, PRO, and NEU messages, along with the language models' sentiment classification decisions on the same messages, are provided in the online supplement (Table S16).

# Results

## GPT-3.5 Evaluation

The average accuracy, based on 20 response instances ($K_{20}$), was .612 ($SE$ = .020) for long-form and .570 ($SE$ = .020) for short-form messages. However, the accuracy varied across categories. For messages categorized as ANTI by human evaluators, $K_{20}$ was .755 ($SE$ = .028) for long-form and .696 ($SE$ = .030) for short-form messages. For messages categorized as PRO by human evaluators, the accuracy was .544 ($SE$ = .033) for long-form and .476 ($SE$ = .033) for short-form messages. The language model's average accuracy for messages classified as NEU by human evaluators was .461 ($SE$ = .040) for long-form and .507 ($SE$ = .042) for short-form messages. Table 1 presents the accuracy of different sentiment labels with varying $m$.

Table 1. Accuracy Varying the Number of Response Instances (GPT-3.5)

| $m$ | Facebook (Long format) | | | | | | | |
|---|---|---|---|---|---|---|---|---|
| | ANTI ($n$ = 200) | | PRO ($n$ = 200) | | NEU ($n$ = 100) | | All ($N$ = 500) | |
| | $K_m$ ($SE$) | $K_m$ / $K_{20}$ | $K_m$ ($SE$) | $K_m$ / $K_{20}$ | $K_m$ ($SE$) | $K_m$ / $K_{20}$ | $K_m$ ($SE$) | $K_m$ / $K_{20}$ |
| 1 | .657 (.023) | 87.1% | .452 (.025) | 83.0% | .428 (.024) | 92.7% | .529 (.015) | 86.5% |
| 3 | .715 (.025) | 94.7% | .506 (.029) | 93.1% | .446 (.030) | 96.7% | .578 (.017) | 94.4% |
| 5 | .734 (.026) | 97.2% | .524 (.030) | 96.2% | .451 (.033) | 97.7% | .593 (.018) | 96.9% |
| 7 | .740 (.027) | 98.0% | .530 (.031) | 97.4% | .455 (.034) | 98.6% | .599 (.018) | 97.9% |
| 9 | .744 (.027) | 98.6% | .535 (.031) | 98.2% | .456 (.036) | 98.7% | .603 (.019) | 98.5% |
| 11 | .749 (.027) | 99.2% | .537 (.032) | 98.7% | .459 (.037) | 99.5% | .606 (.019) | 99.1% |
| 20 | .755 (.028) | - | .544 (.033) | - | .461 (.040) | - | .612 (.020) | - |
| $m$ | Twitter (Short format) | | | | | | | |
| | ANTI ($n$ = 200) | | PRO ($n$ = 200) | | NEU ($n$ = 100) | | All ($N$ = 500) | |
| | $K_m$ ($SE$) | $K_m$ / $K_{20}$ | $K_m$ ($SE$) | $K_m$ / $K_{20}$ | $K_m$ ($SE$) | $K_m$ / $K_{20}$ | $K_m$ ($SE$) | $K_m$ / $K_{20}$ |
| 1 | .614 (.025) | 88.3% | .411 (.025) | 86.4% | .444 (.026) | 87.6% | .499 (.016) | 87.5% |
| 3 | .662 (.027) | 95.1% | .449 (.029) | 94.3% | .481 (.033) | 94.8% | .540 (.018) | 94.8% |
| 5 | .675 (.028) | 97.0% | .459 (.031) | 96.5% | .492 (.036) | 97.1% | .552 (.019) | 96.9% |
| 7 | .681 (.029) | 97.9% | .465 (.031) | 97.6% | .496 (.038) | 97.9% | .557 (.019) | 97.8% |

| 9 | .685 (.029) | 98.4% | .469 (.032) | 98.6% | .502 (.039) | 98.9% | .562 (.019) | 98.6% |
| 11 | .688 (.029) | 99.0% | .470 (.032) | 98.8% | .502 (.040) | 99.0% | .564 (.020) | 98.9% |
| 20 | .696 (.030) | - | .476 (.033) | - | .507 (.042) | - | .570 (.020) | - |

*Note.* $m$ indicates the number of response instances used for majority determination. When $m$ is greater than 1, machine decision is the majority among the response instances. When $m$ equals to 1, the machine decision is equal to the response instance (one-shot determination). $K_m$ indicates the average of the machine accuracy of $n$ messages, when the machine decision of each message was determined based on $m$ response instances. ANTI, PRO, and NEU denote human-evaluated messages expressing anti-HTPs, pro-HTPs, and neutral sentiments, respectively.

Most discrepancies arose when the language model classified messages as NEU or IR, whereas human evaluators identified positive or negative sentiment in these messages. For instance, the model misclassified 24.5% of the human-evaluated long-form ANTI messages. Among these misclassified messages, the language model classified 61.2%, 26.5%, and 12.2% as NEU, IR, and PRO. Table S2, S4 to S7, and Figure S1 and S2 in the online supplement provide more detailed comparative descriptions of decisions made by human evaluators and GPT-3.5.

The significance of differences in accuracy between sentiment categories was examined. The results indicated that $K_{20}$ for human-evaluated ANTI messages was significantly higher than that of human-labeled PRO ($U$ = 26483.5, $p$ < .001) and NEU ($U$ = 14675.5, $p$ < .001) messages in long-form. This was also true for human-labeled short-form PRO ($U$ = 25876.5, $p$ < .001) and NEU ($U$ = 13312.5, $p$ < .001) messages. These gaps in accuracy were consistent across all $m$ values and formats (see Table S12 and S13 in the online supplement).

The accuracy improved as $m$ increased, as visualized in Figure 1. However, even with a few response instances, the accuracy was comparable to the accuracy based on 20 response instances. For example, even the average accuracy of one-shot determination ($K_1$) for human-labeled ANTI, PRO, and NEU messages also reached 87.1%, 83.0%, and 92.7% of $K_{20}$ for long-form messages and 88.3%, 86.4%, and 87.6% of $K_{20}$ for short-form messages.

Figure 1. Accuracy Across Response Instances and Message Formats (GPT-3.5)

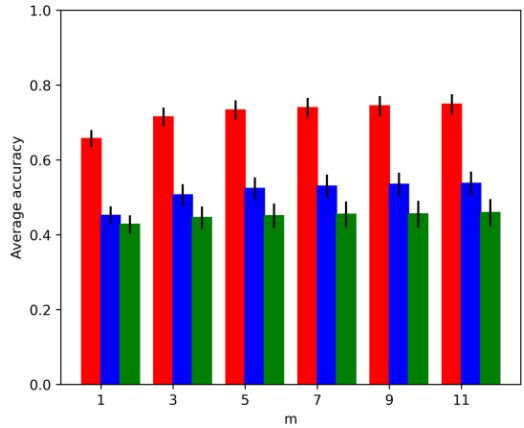
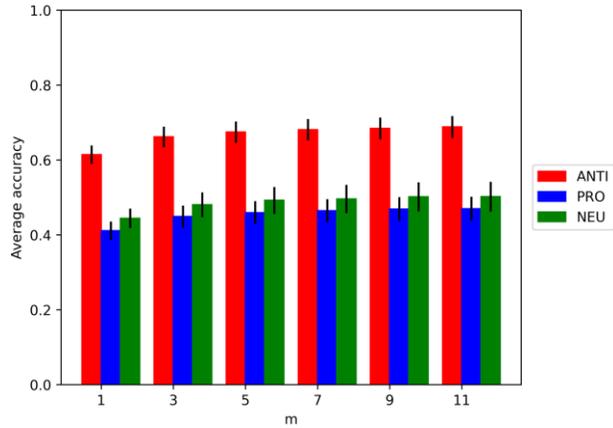

*Note.* Error bars represent mean ± standard error of the mean. "m" refers to the number of response instances.

### GPT-4 Turbo Evaluation

GPT-4 Turbo demonstrated higher accuracy than GPT-3.5 across all sentiment categories. The language model's overall average accuracy was .817 (*SE* = .017) for long-form messages and .770 (*SE* = .019) for short-form messages. Although accuracy varied across categories, the gap between the accuracy in ANTI and PRO sentiment classification decreased compared to GPT-3.5. For human-labeled ANTI messages, $K_{20}$ was .861 (*SE* = .024) for long-form messages and .790 (*SE* = .028) for short-form messages. For human-labeled PRO messages, $K_{20}$ was .840 (*SE* = .025) for long-form messages and .783 (*SE* = .029) for short-form messages. The accuracy for human-labeled NEU message categorization also increased compared to GPT-3.5. For NEU messages, $K_{20}$ was .685 (*SE* = .044) for long-form messages and .703 (*SE* = .045) for short-form messages. The accuracy of sentiment categories with varying *m* is reported in Table 2.

Table 2. Accuracy Varying the Number of Response Instances (GPT-4 Turbo)

| | Facebook (Long format) | | | | | | | |
|---|---|---|---|---|---|---|---|---|
| *m* | ANTI (*n* = 200) | | PRO (*n* = 200) | | NEU (*n* = 100) | | All (*N* = 500) | |
| | $K_m$ (*SE*) | $K_m$ / $K_{20}$ | $K_m$ (*SE*) | $K_m$ / $K_{20}$ | $K_m$ (*SE*) | $K_m$ / $K_{20}$ | $K_m$ (*SE*) | $K_m$ / $K_{20}$ |
| 1 | .856 (.023) | 99.4% | .828 (.024) | 98.5% | .661 (.038) | 96.5% | .806 (.016) | 98.6% |
| 3 | .859 (.024) | 99.7% | .834 (.025) | 99.3% | .681 (.041) | 99.3% | .813 (.016) | 99.5% |
| 5 | .860 (.024) | 99.9% | .837 (.025) | 99.6% | .685 (.042) | 100% | .816 (.016) | 99.8% |
| 7 | .860 (.024) | 99.9% | .837 (.025) | 99.6% | .687 (.042) | 100% | .816 (.017) | 99.8% |
| 9 | .860 (.024) | 99.9% | .838 (.025) | 99.7% | .687 (.043) | 100% | .816 (.017) | 99.9% |
| 11 | .860 (.024) | 99.9% | .838 (.025) | 99.8% | .688 (.043) | 100% | .817 (.017) | 100% |
| 20 | .861 (.024) | - | .840 (.025) | - | .685 (.044) | - | .817 (.017) | - |
| | Twitter (Short format) | | | | | | | |
| $m$ | ANTI ($n$ = 200) | | PRO ($n$ = 200) | | NEU ($n$ = 100) | | All ($N$ = 500) | |
| | $K_m$ (SE) | $K_m$ / $K_{20}$ | $K_m$ (SE) | $K_m$ / $K_{20}$ | $K_m$ (SE) | $K_m$ / $K_{20}$ | $K_m$ (SE) | $K_m$ / $K_{20}$ |
| 1 | .789 (.027) | 99.9% | .773 (.028) | 98.7% | .704 (.040) | 100% | .765 (.018) | 99.4% |
| 3 | .789 (.028) | 99.9% | .776 (.028) | 99.2% | .708 (.042) | 100% | .768 (.018) | 99.8% |
| 5 | .789 (.028) | 99.9% | .778 (.028) | 99.4% | .707 (.043) | 100% | .768 (.018) | 99.8% |
| 7 | .790 (.028) | 100% | .779 (.028) | 99.6% | .706 (.043) | 100% | .769 (.018) | 99.9% |
| 9 | .789 (.028) | 99.9% | .780 (.029) | 99.7% | .705 (.044) | 100% | .769 (.018) | 99.9% |
| 11 | .789 (.028) | 99.9% | .781 (.029) | 99.7% | .705 (.044) | 100% | .769 (.018) | 99.9% |
| 20 | .790 (.028) | - | .783 (.029) | - | .703 (.045) | - | .770 (.019) | - |

*Note.* $m$ indicates the number of response instances used for majority determination. When $m$ is greater than 1, machine decision is the majority among the response instances. When $m$ equals to 1, the machine decision is equal to the response instance (one-shot determination). $K_m$ indicates the average of the machine accuracy of $n$ messages, when the machine decision of each message was determined based on $m$ response instances. ANTI, PRO, and NEU denote human-evaluated messages expressing anti-HTPs, pro-HTPs, and neutral sentiments, respectively.

GPT-4 Turbo showed fewer false selections of the IR label across all sentiment categories. When examined with three randomly selected response instances ($m$ = 3), the model misclassified only 10 out of the entire 1,000 sample messages as IR. This impacted the pattern of mismatches between human and language model sentiment classification. For instance, the model categorized 14.2% of human-labeled ANTI messages as one of the other labels. Of these mismatches, 91.8% were classified as NEU whereas only 0.09% were IR. The online supplement includes contingency tables (Table S3, and S8 to S11) as well as flow charts (Figure S2) providing more detailed comparative descriptions of human and language model sentiment labeling.

For long-form messages, the model's accuracy of ANTI classification based on 20 response instances was significantly higher than NEU classification ($U$ = 12561.5, $p$ < .001) but not significantly different from the accuracy of PRO classification ($U$ = 20810.5, $p$ = .30). PRO classification also showed significantly higher accuracy compared to NEU classification ($U$ = 12176, $p$ < .001). This pattern was observed across all *m* values. For short-form messages, with lower m values, the accuracy of ANTI and PRO classification was significantly greater than NEU labeling. For example, the one-shot determination ($K_1$) for ANTI and PRO classification was significantly higher than for NEU classification (ANTI vs. NEU: $U = 11829$, $p < .001$; PRO vs. NEU: $U = 11625$, $p < .001$). However, these differences diminished as *m* increased; $K_{20}$ was not significantly different across the three sentiment categories. The difference test results across all *m* values are provided in the online supplement (Table S14 and S15).

The model's accuracy in one-shot cases was already comparable to that of 20 instances, as shown in Figure 2. Table 2 above presented that the accuracy of the one-shot determination ($K_1$) reached at least 96.5% of the accuracy based on 20 response instances ($K_{20}$).

Figure 2. Accuracy Across Response Instances and Message Formats (GPT-4 Turbo)

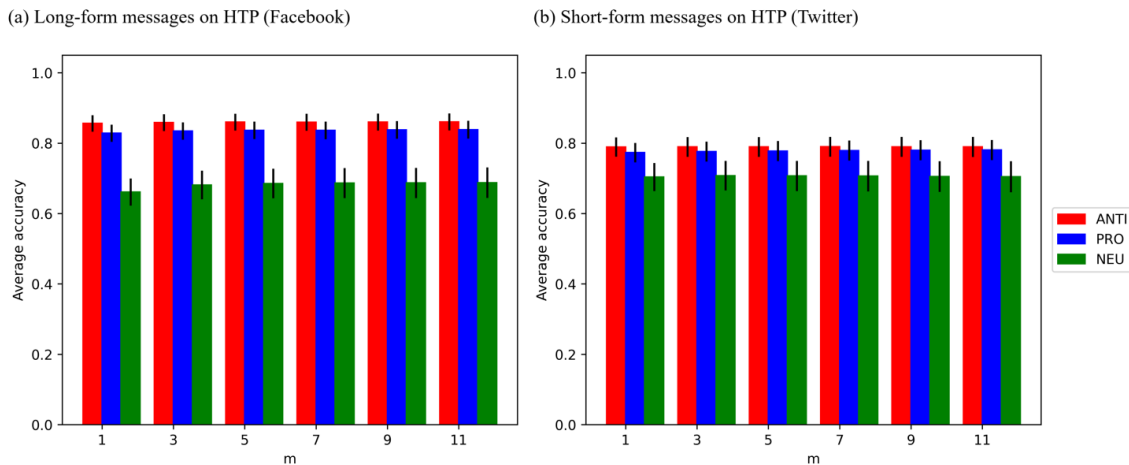

*Note.* Error bars represent mean ± standard error of the mean. "m" refers to the number of response instances.

## Discussion

GPT-4 Turbo accurately replicated 81.7% of human sentiment evaluations for long-form messages and 77.0% for short-form messages, based on 20 AI responses. In comparison, GPT-3.5's $K_{20}$ indicated that the model's labeling decisions matched human coders' evaluations with 61.2% accuracy for long-form messages and 57.0% accuracy for short-form messages. In sum, GPT-4 Turbo showed improvements in accuracy compared to GPT-3.5, due to increased accuracy across all sentiment categories.

Focusing on GPT-4 Turbo, which showed better overall accuracy, the LLM demonstrated already high accuracy with a small number of responses. The difference in accuracy between a small number of responses (e.g. $m$ = 1, 3) and a high number of responses (e.g. $m$ = 20) was not statistically significant. For all sentiment categories, the language model's $K_3$ reached at least 99.2% of $K_{20}$. The model demonstrated similar levels of accuracy for ANTI and PRO labels for both long-form and short-form messages. While the accuracy of NEU classification was lower than that of ANTI and PRO classification, it increased by approximately 20 percentage points compared to GPT-3.5. These findings suggest that GPT-4 Turbo can yield more accurate sentiment classification decisions, even with a small number of response instances, such as three.

GPT-3.5 showed discrepancies in accuracy across sentiment categories. Specifically, the accuracy of ANTI classification was better than PRO classification. This finding suggests the possibility of a relative under-representation of messages with positive sentiment compared to negative sentiment when utilizing LLMs for sentiment analysis of tobacco-related social media discourses. This issue calls for further exploration of approaches, techniques, and procedures to assess, mitigate, or compensate for LLMs' potential inconsistencies across different sentiment categories, as well as reasons underlying these discrepancies.

Employing a newer model may be the most straightforward solution, as shown by the comparison between older (GPT-3.5) and newer (GPT-4 Turbo) models. The newer model not only improved accuracy across all sentiment categories but also showed no significant difference in accuracy between ANTI and PRO classification. This is particularly important for tobacco prevention researchers, as the detection of ANTI and PRO sentiments are important due to their possible associations with tobacco prevention outcomes [11,12,31]. However, a more recent model may not always perform better than its predecessors. For example, GPT-4 Turbo experienced a "laziness" issue, where the model does not complete user requests [32]. Therefore, performance of new language models on specific tasks should also be rigorously tested.

Utilizing language models specialized for health and medical information analyses such as Google's Med-PaLM [33] and Stanford University's BioMedLM (previously PubMedGPT) [34] may influence accuracy. Research in this area is still emerging, with limited evidence on the application of these specialized LLMs for sentiment analysis. Previous studies have primarily utilized standard GPT models [35–38]. Also, using these specialized models could be more difficult than widely used platforms like ChatGPT. Still, their performance in analyzing sentiment of public health-related social media messages warrants future investigations, considering their capabilities in handling content from the general public, not just academic researchers and professionals.

Prompt engineering could be another strategy for improving the accuracy of LLMs in sentiment analysis and reducing discrepancies across categories. In line with the objective of this study to facilitate a straightforward comparison between human coders' and language models' labeling decisions, we used a prompt that closely mirrored the coding scheme for human evaluators. However, different prompting techniques can lead to different results for similar requests [39,40]. Techniques such as few-shot prompting, which involves including task-related examples within the prompts, may enhance accuracy. For instance, rather than only defining sentiment labels, the coding scheme can provide several example messages for each label. Although these techniques complicate direct comparisons between human and machine classification, they possibly offer potential accuracy gains worth exploring.

Establishing and adopting coding procedures for LLM-involved coding is also worthwhile. A study investigating LLMs as substitutes for human coders in labeling texts on political topics serves as a good example [30]. The authors proposed a 'hybrid' model where disagreements between the 'GPT-4 first run' and the 'GPT-4 second run' are resolved by a human coder. Their findings demonstrated that the hybrid approach can be optimized with minimal additional human effort and boosted the accuracy of GPT-4's annotations. This hybrid approach can potentially be adapted for the analysis of sentiment about health topics, and other coding procedures should be explored to further enhance the accuracy and efficiency.

The implications and future applications of our findings should be discussed with caution. Firstly, the current study is a focused case study on OpenAI's GPT, examining sentiment analysis on Twitter and Facebook messages related to HTPs. Future research can extend beyond this specific focus to evaluate the accuracy of LLMs in analyzing the sentiment of health-related information across a broader range of topics and platforms. Secondly, while the differences in accuracy for ANTI and PRO sentiment classifications that were present in GPT-3.5 disappeared in GPT-4 Turbo, and the accuracy of NEU classification increased by 20%, the NEU classification accuracy remains lower. A simple explanation might be the underperformance of the models. For instance, GPT-4 Turbo misclassified a human-labeled NEU message, which used HTPs as examples to explain an economic principle, as irrelevant. Or, this difference might stem from the inherent complexity in evaluating neutrality [15,41]. For example, human coders classified a message as NEU, describing IQOS as a device that uses a "patented heat-control technology." In contrast, GPT-4 Turbo classified the same message as PRO, interpreting "patented heat-control technology" to have positive connotations. It may be worthwhile for future research to explore the patterns of misclassifications by LLMs. Thirdly, this study did not address potential ethical issues of utilizing LLMs for sentiment analysis of social media content. Ethical considerations, such as security, privacy protection, and data ownership, are important when using LLMs to analyze social media messages [42,43]. These issues require careful attention when analyzing health-related information using LLMs, too [44,45]. Future research should utilize LLMs while carefully considering the potential ethical issues surrounding their content analyses.

## Conflicts of Interest

The authors declare no conflict of interest.

## Acknowledgement

We thank Kyungmin Kang for her contribution in message evaluation.

## Data Availability

The data sets generated during and/or analyzed during this study are available in the Open Science Framework (OSF) repository: https://doi.org/10.17605/osf.io/6teuz

# Abbreviations

AI: Artificial Intelligence

CT: CrowdTangle

GPT: Generative Pre-trained Transformer

HTPs: Heated tobacco products

LLMs: Large Language Models

Online Supplements for "Large Language Models' Accuracy in Emulating Human Experts' Evaluation of Public Sentiments about Heated Tobacco Products on Social Media: Evaluation Study"

Kwanho Kim, PhD[1] & Soojong Kim, PhD[2]

[1] Department of Media, College of Politics and Economics, Kyung Hee University, Korea

[2] Department of Communication, University of California Davis, United States

**Author Note**

Soojong Kim  https://orcid.org/0000-0002-1334-5310

Correspondence concerning this article should be addressed to Soojong Kim, 1 Shields Ave, Kerr Hall #361, Davis, CA 95616, United States. Email: sjokim@ucdavis.edu

**Human Evaluation Dataset Generation Process**

The human-annotated datasets were generated in the following four steps. The multi-step random sampling procedure was designed to enhance the likelihood of including messages with potentially positive and negative sentiments, ensuring their inclusion in the selected messages for the human annotation. For short-form messages, first, from the entire message pool of 60,031 tweets, 200 messages were randomly sampled. Two human evaluators independently classified these messages into one of the following five categories: ANTI (anti-HTPs messages), PRO (pro-HTPs messages), NEU (neutral messages), MIX (messages containing a mixture of positive and negative attitudes on HTPs), and IR (messages not relevant to HTPs). After completing categorizations, evaluators checked differences in categorization decisions between evaluators and resolved these discrepancies via discussion. Second, using these 200 human-evaluated messages, we developed preliminary machine-learning ANTI and PRO binary classifiers to analyze the sentiment of all messages in the entire dataset. These classifiers were developed using DistilBERT, a widely-used pre-trained machine-learning model [1]. Third, using these preliminary machine-classification results, we randomly sampled an additional 500 ANTI and 500 PRO messages from the pool. Lastly, a team of three evaluators analyzed the sentiment of these 1,000 messages. Messages were first evaluated independently by two evaluators. The role of a third evaluator was to resolve disagreements in sentiment classifications between these two initial evaluators. Each initial evaluator was either a postdoctoral-level public health communication scientist or an undergraduate student who had received specialized training and exercise for this sentiment analysis task. The third evaluator was a postdoctoral-level scholar specializing in social media and health communication research. The inter-coder reliability of the evaluators, measured with Cohen's Kappa, was .885.

For long-form samples, 1,250 samples were prepared using a similar procedure with a few additional steps. The first and second steps were identical to those for short-form samples. 200 messages were randomly sampled from the entire message pool of 16,284 Facebook posts and then analyzed for sentiment by two independent human evaluators to prepare training materials for preliminary classifiers. However, the preliminary classification results showed that there were only 144 PRO messages in the entire long-form message pool. Prior to the next step, preliminary classifications were re-conducted using improved preliminary classifiers re-trained with 250 samples. These 250 samples were prepared by adding 50 more human-evaluated random samples to the 200 samples that had already been evaluated by human evaluators for sentiment. Using these re-trained preliminary classifiers, the sentiment of all long-form messages was analyzed. Identical to the short-form message sampling procedure, 500 ANTI and 500 PRO messages were randomly sampled from the entire dataset using preliminary classification results from the re-trained classifiers. Then, these 1,000 messages were evaluated by three human evaluators, as they did for short-form samples. The reliability measure, represented by Cohen's Kappa, was .938.

Table S1. Prompt for Machine Evaluation

| Prompt |
| --- |
| The content presented in [Message] is a post on {Platform}. [Message] contains content generated or re-shared by an author. What does the author of the post argue? Classify the post into one of the categories explained in [Scheme]. Start your response with a category code you choose (IR, PRO, ANTI, MIX, NEU) without any preceding characters or information before it. Then, print a colon ': ' following the code, and proceed to explain your decision.<br><br>[Message]<br>{Content}<br><br>[Scheme]<br>Below, HTP stands for heated tobacco products, also known as HNB (heat-not-burn) tobacco products or tobacco heating products. HTP brands include IQOS, Glo, Eclipse, and Ploom.<br>1.Irrelevant messages (IR): Use this label if a message is not related to HTP.<br>2.Pro-HTP messages (PRO): Use this label if a message is primarily about HTP and clearly supportive of the use of HTP, the HTP industry, and policies allowing and promoting the use of HTP.<br>3.Anti-HTP messages (ANTI): Use this label if a message is primarily about HTP and clearly against the use of HTP, the HTP industry, and policies preventing and discouraging the use of HTP.<br>4.Mixed messages (MIX): Use this label if a message is primarily about HTP and contains balanced information that both supports and opposes HTP.<br>5.Neutral or 'Not Applicable' messages (NEU): Use this label if a message meets one or more of the following criteria: (a) the message is not primarily about HTP, (b) the length of the message is too short to understand its meaning, or (c) the message neither supports nor opposes the use of HTP, the HTP industry, and policies influencing the use of HTP. |
| Example response instances |
| - "ANTI: The author of the post is arguing against the use of heated tobacco products (HTP) by stating that they contain nicotine and emphasizing the need for more research to understand the health effects of these products."<br>- "PRO: The author of the post argues in favor of the proposed sin tax reform, which aims to raise the excise tax rate on alcohol, heated tobacco products (HTPs), and vaping products." |

*Note*. The placeholder {Platform} was substituted with either "Facebook" or "Twitter", while {Content} was substituted with the actual content of a social media post.

# Comparative Cross-Tabulation of Human and GPT Sentiment Evaluations

Due to the probabilistic nature of LLMs, a message's category determined by GPT models can vary across different response instances. Considering these variations, Table S2 and S3 were calculated based on the following procedure. First, for each message, three out of 20 response instances were randomly selected with replacement. Second, the majority sentiment among the three selected instances was determined. Third, the first and second processes were conducted for all messages, and a cross-tabulation result was calculated by comparing the human- and machine-evaluations of the messages. We repeated these processes 1,000 times, and then the 1,000 cross-tabulation results generated from these iterations were averaged.

Table S2. Comparative Cross-Tabulation of Human and GPT-3.5 Sentiment Evaluations

| | Facebook (Long format) | | | |
|---|---|---|---|---|
| GPT-3.5 | Human | | | |
| | ANTI | PRO | NEU | Total |
| ANTI | 151.123 (.084) 75.6% | 13.691 (.055) 6.8% | 8.987 (.051) 9.0% | 173.801 34.8% |
| PRO | 5.788 (.037) 2.9% | 109.005 (.085) 54.5% | 8.641 (.044) 8.6% | 123.434 24.7% |
| NEU | 30.139 (.081) 15.1% | 29.771 (.093) 14.9% | 46.060 (.093) 46.1% | 105.970 21.2% |
| MIX | 0.020 (.005) 0.01% | 0.529 (.021) 0.2% | 0.465 (.014) 0.5% | 1.014 0.2% |
| IR | 12.930 (.075) 6.5% | 47.068 (.099) 23.5% | 35.783 (.089) 35.8% | 95.781 19.1% |
| Total | 200 100% | 200 100% | 100 100% | 500 100% |
| | Twitter (Short format) | | | |
| GPT-3.5 | Human | | | |
| | ANTI | PRO | NEU | Total |
| ANTI | 139.130 (.083) 69.6% | 27.454 (.064) 13.7% | 12.754 (.051) 12.8% | 179.338 35.9% |
| PRO | 4.757 (.045) 2.4% | 95.225 (.081) 47.6% | 9.278 (.053) 9.3% | 109.260 21.9% |
| NEU | 40.220 (.092) 20.1% | 48.818 (.101) 24.4% | 50.647 (.085) 50.6% | 139.685 27.9% |
| MIX | 0.411 (.017) 0.2% | 0.842 (.027) 0.4% | 1.882 (.032) 1.9% | 3.135 0.6% |
| IR | 15.482 (.080) 7.7% | 27.661 (.096) 13.8% | 25.439 (.081) 25.4% | 68.582 13.7% |
| Total | 200 (100%) | 200 (100%) | 100 (100%) | 500 (100%) |

*Note*. The table includes counts and standard errors (noted in parentheses), along with column proportions.

Table S3. Comparative Cross-Tabulation of Human and GPT-4 Turbo Sentiment Evaluations

| Facebook (Long format) | | | | |
|---|---|---|---|---|
| GPT-4 Turbo | Human | | | |
|  | ANTI | PRO | NEU | Total |
| ANTI | 171.708 (.040) 85.9% | 3.029 (.008) 1.5% | 11.321 (.024) 11.3% | 186.058 37.2% |
| PRO | 0.233 (.014) 0.1% | 166.906 (.056) 83.5% | 15.148 (.054) 15.1% | 182.287 36.5% |
| NEU | 25.963 (.039) 13.0% | 26.401 (.058) 13.2% | 68.097 (.040) 68.1% | 120.461 24.1% |
| MIX | 2.071 (.026) 1.0% | 2.354 (.017) 1.2% | 1.694 (.029) 1.7% | 6.119 1.2% |
| IR | 0.025 (.005) 0.0% | 1.31 (.031) 0.7% | 3.74 (.044) 3.7% | 5.075 1.0% |
| Total | 200 100% | 200 100% | 100 100% | 500 100% |
| Twitter (Short format) | | | | |
| GPT-4 Turbo | Human | | | |
|  | ANTI | PRO | NEU | Total |
| ANTI | 157.789 (.047) 78.9% | 5.311 (.021) 2.7% | 8.69 (.030) 8.7% | 171.79 34.4% |
| PRO | 4.233 (.020) 2.1% | 155.363 (.058) 77.7% | 15.569 (.035) 15.6% | 175.165 35.0% |
| NEU | 32.963 (.054) 16.5% | 32.031 (.052) 16.0% | 70.721 (.059) 70.7% | 135.715 27.1% |
| MIX | 4.383 (.027) 2.2% | 7.148 (.033) 3.6% | 1.166 (.021) 1.2% | 12.697 2.5% |
| IR | 0.632 (.020) 0.3% | 0.147 (.011) 0.1% | 3.854 (.028) 3.9% | 4.633 0.9% |
| Total | 200 (100%) | 200 (100%) | 100 (100%) | 500 (100%) |

*Note*. The table includes counts and standard errors (noted in parentheses), along with column proportions.

**Binary Confusion Matrix of Human and GPT Models' Sentiment Evaluations**

In this study, we transformed the multi-category sentiment classifications made by human evaluators and GPT models into binary classifications to generate confusion matrices for ANTI and PRO classifications. For the ANTI classifications, we grouped PRO, NEU, MIX, and IR as 'Non-ANTI,' to generate binary confusion matrices. Similarly, for the PRO classifications, we grouped ANTI, NEU, MIX, and IR as 'Non-PRO,' to create binary confusion matrices. The data used for these comparisons were derived from the cross-tabulation analyses results reported in Table S2 and S3.

Table S4. Binary Confusion Matrix of ANTI Classification of Facebook Messages (GPT-3.5)

|  |  | **Human Coders' Evaluation** | |
|---|---|---|---|
|  |  | ANTI | Non-ANTI |
| **GPT-3.5's Evaluation** | ANTI | 151.123 | 22.678 |
|  | Non-ANTI | 48.877 | 277.322 |

*Note.*
True Positive Rate (Sensitivity) = .756
False Positive Rate (False Alarm) = .076
True Negative Rate (Specificity) = .924
False Negative Rate (Miss) = .244

Precision = .869
Recall = .756
F1 = .808

Table S5. Binary Confusion Matrix of PRO Classification of Facebook Messages (GPT-3.5)

|  |  | **Human Coders' Evaluation** | |
|---|---|---|---|
|  |  | PRO | Non-PRO |
| **GPT-3.5's Evaluation** | PRO | 109.005 | 14.429 |
|  | Non-PRO | 90.995 | 285.571 |

*Note.*
True Positive Rate (Sensitivity) = .545
False Positive Rate (False Alarm) = .048
True Negative Rate (Specificity) = .952
False Negative Rate (Miss) = .455

Precision = .883
Recall = .545
F1 = .674

Table S6. Binary Confusion Matrix of ANTI Classification of Twitter Messages (GPT-3.5)

|  |  | **Human Coders' Evaluation** | |
|---|---|---|---|
|  |  | ANTI | Non-ANTI |
| **GPT-3.5's Evaluation** | ANTI | 139.13 | 40.208 |
|  | Non-ANTI | 60.87 | 259.792 |

*Note.*
True Positive Rate (Sensitivity) = .696
False Positive Rate (False Alarm) = .134
True Negative Rate (Specificity) = .866
False Negative Rate (Miss) = .304

Precision = .776
Recall = .696
F1 = .734

Table S7. Binary Confusion Matrix of PRO Classification of Twitter Messages (GPT-3.5)

|  |  | **Human Coders' Evaluation** | |
|---|---|---|---|
|  |  | PRO | Non-PRO |
| **GPT-3.5's Evaluation** | PRO | 95.225 | 14.035 |
|  | Non-PRO | 104.775 | 285.965 |

*Note.*
True Positive Rate (Sensitivity) = .476
False Positive Rate (False Alarm) = .047
True Negative Rate (Specificity) = .953
False Negative Rate (Miss) = .524

Precision = .872
Recall = .476
F1 = .617

Table S8. Binary Confusion Matrix of ANTI Classification of Facebook Messages (GPT-4 Turbo)

|  |  | Human Coders' Evaluation | |
| --- | --- | --- | --- |
|  |  | ANTI | Non-ANTI |
| **GPT-4 Turbo's Evaluation** | ANTI | 171.708 | 14.35 |
|  | Non-ANTI | 28.292 | 285.65 |
| *Note.*  True Positive Rate (Sensitivity) = .858  False Positive Rate (False Alarm) = .048  True Negative Rate (Specificity) = .952  False Negative Rate (Miss) = .142  Precision = .923  Recall = .858  F1 = .889 | | | |

Table S9. Binary Confusion Matrix of PRO Classification of Facebook Messages (GPT-4 Turbo)

|  |  | Human Coders' Evaluation | |
| --- | --- | --- | --- |
|  |  | PRO | Non-PRO |
| **GPT-4 Turbo's Evaluation** | PRO | 166.906 | 15.381 |
|  | Non-PRO | 33.094 | 284.619 |
| *Note.*  True Positive Rate (Sensitivity) = .834  False Positive Rate (False Alarm) = .051  True Negative Rate (Specificity) = .949  False Negative Rate (Miss) = .166  Precision = .916  Recall = .834  F1 = .873 | | | |

Table S10. Binary Confusion Matrix of ANTI Classification of Twitter Messages (GPT-4 Turbo)

|  |  | **Human Coders' Evaluation** | |
|---|---|---|---|
|  |  | ANTI | Non-ANTI |
| **GPT-4 Turbo's Evaluation** | ANTI | 157.789 | 14.048 |
|  | Non-ANTI | 42.211 | 286.119 |

*Note.*
True Positive Rate (Sensitivity) = .789
False Positive Rate (False Alarm) = .047
True Negative Rate (Specificity) = .953
False Negative Rate (Miss) = .211

Precision = .918
Recall = .789
F1 = .849

Table S11. Binary Confusion Matrix of PRO Classification of Twitter Messages (GPT-4 Turbo)

|  |  | **Human Coders' Evaluation** | |
|---|---|---|---|
|  |  | PRO | Non-PRO |
| **GPT-4 Turbo's Evaluation** | PRO | 155.363 | 19.802 |
|  | Non-PRO | 44.637 | 280.198 |

*Note.*
True Positive Rate (Sensitivity) = .777
False Positive Rate (False Alarm) = .066
True Negative Rate (Specificity) = .934
False Negative Rate (Miss) = .223

Precision = .887
Recall = .777
F1 = .828

Figure S1. Comparative Visualizations of Sentiment Evaluations by Humans and GPT-3.5 on the Same Messages

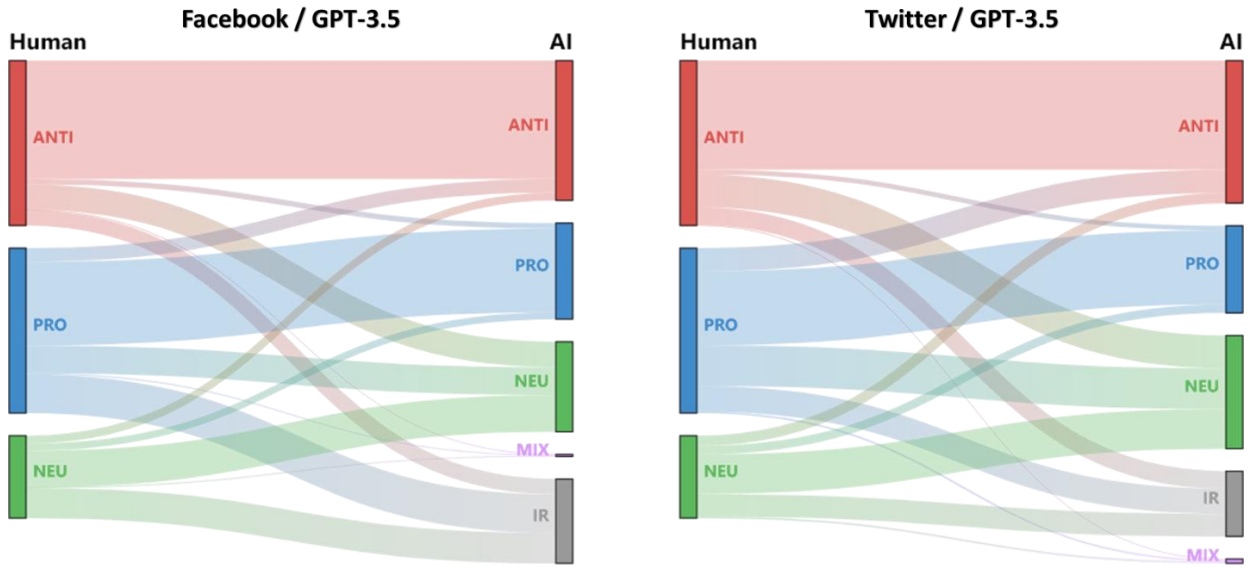

*Note.* The diagrams provide comparative illustrations of the sentiment labeling decisions on the same set of Facebook and Twitter messages, made by human evaluators and GPT-3.5.

Figure S2. Comparative Visualizations of Sentiment Evaluations by Humans and GPT-4 Turbo on the Same Messages

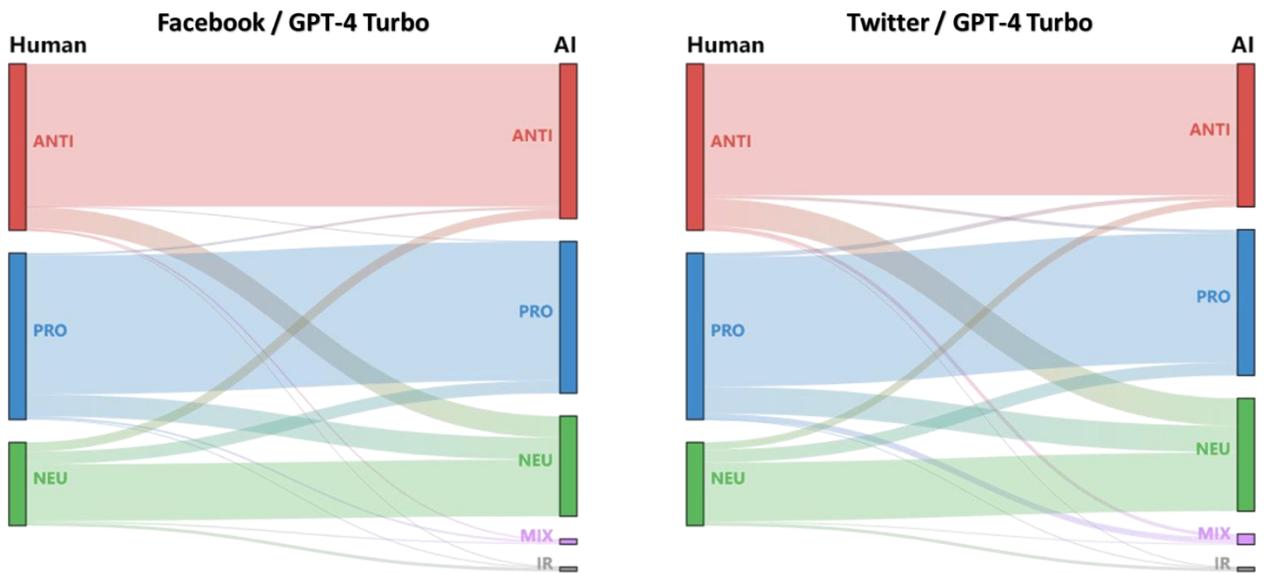

*Note*. The diagrams provide comparative illustrations of the sentiment labeling decisions on the same set of Facebook and Twitter messages, made by human evaluators and GPT-4 Turbo.

Table S12. Difference in Average Accuracy Between Evaluation Sets (Facebook / GPT-3.5)

| $m$ | Eval Set 1 | Eval Set 2 | Mann-Whitney $U$ | $p$ |
|---|---|---|---|---|
| 1 | ANTI | PRO | 26988.5 | <.001 |
| 1 | ANTI | NEU | 14390.5 | <.001 |
| 1 | PRO | NEU | 10240 | .74 |
| 3 | ANTI | PRO | 26784.5 | <.001 |
| 3 | ANTI | NEU | 14697 | <.001 |
| 3 | PRO | NEU | 10709 | .32 |
| 5 | ANTI | PRO | 26587 | <.001 |
| 5 | ANTI | NEU | 14837.5 | <.001 |
| 5 | PRO | NEU | 10857 | .23 |
| 7 | ANTI | PRO | 26697 | <.001 |
| 7 | ANTI | NEU | 14854.5 | <.001 |
| 7 | PRO | NEU | 10930.5 | .19 |
| 9 | ANTI | PRO | 26594.5 | <.001 |
| 9 | ANTI | NEU | 14823 | <.001 |
| 9 | PRO | NEU | 10945.5 | .18 |
| 11 | ANTI | PRO | 26357 | <.001 |
| 11 | ANTI | NEU | 14821 | <.001 |
| 11 | PRO | NEU | 11029.5 | .14 |
| 20 | ANTI | PRO | 26483.5 | <.001 |
| 20 | ANTI | NEU | 14675.5 | <.001 |
| 20 | PRO | NEU | 10987.5 | .16 |

Table S13. Difference in Average Accuracy Between Evaluation Sets (Twitter / GPT-3.5)

| $m$ | Eval Set 1 | Eval Set 2 | Mann-Whitney $U$ | $p$ |
|---|---|---|---|---|
| 1 | ANTI | PRO | 26866 | <.001 |
| 1 | ANTI | NEU | 13192.5 | <.001 |
| 1 | PRO | NEU | 9088.5 | .20 |
| 3 | ANTI | PRO | 26827.5 | <.001 |
| 3 | ANTI | NEU | 13425 | <.001 |
| 3 | PRO | NEU | 9357 | .36 |
| 5 | ANTI | PRO | 26760 | <.001 |
| 5 | ANTI | NEU | 13423 | <.001 |
| 5 | PRO | NEU | 9345 | .36 |
| 7 | ANTI | PRO | 26444 | <.001 |
| 7 | ANTI | NEU | 13402 | <.001 |
| 7 | PRO | NEU | 9440 | .43 |
| 9 | ANTI | PRO | 26466.5 | <.001 |
| 9 | ANTI | NEU | 13322 | <.001 |
| 9 | PRO | NEU | 9413.5 | .41 |
| 11 | ANTI | PRO | 26243.5 | <.001 |
| 11 | ANTI | NEU | 13374.5 | <.001 |
| 11 | PRO | NEU | 9500 | .48 |
| 20 | ANTI | PRO | 25876.5 | <.001 |
| 20 | ANTI | NEU | 13312.5 | <.001 |
| 20 | PRO | NEU | 9603.5 | .57 |

Table S14. Difference in Average Accuracy Between Evaluation Sets (Facebook / GPT-4 Turbo)

| *m* | Eval Set 1 | Eval Set 2 | Mann-Whitney *U* | *p* |
|---|---|---|---|---|
| 1 | ANTI | PRO | 21271.5 | .15 |
| 1 | ANTI | NEU | 14308 | < .001 |
| 1 | PRO | NEU | 13674 | < .001 |
| 3 | ANTI | PRO | 21270 | .15 |
| 3 | ANTI | NEU | 14281.5 | < .001 |
| 3 | PRO | NEU | 13654.5 | < .001 |
| 5 | ANTI | PRO | 21338 | .12 |
| 5 | ANTI | NEU | 14190 | < .001 |
| 5 | PRO | NEU | 13549.5 | < .001 |
| 7 | ANTI | PRO | 21186.5 | .16 |
| 7 | ANTI | NEU | 13506.5 | < .001 |
| 7 | PRO | NEU | 12938.5 | < .001 |
| 9 | ANTI | PRO | 20926 | .26 |
| 9 | ANTI | NEU | 13269.5 | < .001 |
| 9 | PRO | NEU | 12796 | < .001 |
| 11 | ANTI | PRO | 20926.5 | .25 |
| 11 | ANTI | NEU | 13158.5 | < .001 |
| 11 | PRO | NEU | 12692.5 | < .001 |
| 20 | ANTI | PRO | 20810.5 | .30 |
| 20 | ANTI | NEU | 12561.5 | < .001 |
| 20 | PRO | NEU | 12176 | < .001 |

Table S15. Difference in Average Accuracy Between Evaluation Sets (Twitter / GPT-4 Turbo)

| $m$ | Eval Set 1 | Eval Set 2 | Mann-Whitney $U$ | $p$ |
|---|---|---|---|---|
| 1 | ANTI | PRO | 20222 | .81 |
| 1 | ANTI | NEU | 11829 | .003 |
| 1 | PRO | NEU | 11625 | .008 |
| 3 | ANTI | PRO | 20218 | .82 |
| 3 | ANTI | NEU | 11812 | .003 |
| 3 | PRO | NEU | 11626 | .008 |
| 5 | ANTI | PRO | 20152 | .87 |
| 5 | ANTI | NEU | 11596 | .008 |
| 5 | PRO | NEU | 11431.5 | .02 |
| 7 | ANTI | PRO | 20496 | .56 |
| 7 | ANTI | NEU | 11494.5 | .011 |
| 7 | PRO | NEU | 11212.5 | .04 |
| 9 | ANTI | PRO | 20673.5 | .45 |
| 9 | ANTI | NEU | 11562.5 | .007 |
| 9 | PRO | NEU | 11223 | .04 |
| 11 | ANTI | PRO | 20828.5 | .35 |
| 11 | ANTI | NEU | 11235.5 | .03 |
| 11 | PRO | NEU | 10837 | .15 |
| 20 | ANTI | PRO | 20385 | .66 |
| 20 | ANTI | NEU | 10932 | .09 |
| 20 | PRO | NEU | 10744.5 | .18 |

Table S16. Human and GPT-4 Turbo (GPT-4T) Evaluation Examples

| Facebook Messages | Human | GPT-4T |
|---|---|---|
| Big Tobacco's new cigarette is sleek, smokeless — but is it any better for you? Experts doubt motive, science behind new heated tobacco tech. FDA will decide soon whether to allow its sale. Tobacco companies have a long history of lying about their deadly products, so any claims about safer tobacco products must be regarded with skepticism. To protect public health, we need strong FDA regulation of all tobacco products and any health claims about them. | ANTI | ANTI |
| The IQOS system is an elegant and innovative alternative to smoking using tobacco instead of Eliquids. IQOS doesn't burn the tobacco it heats it for the nicotine and real tobacco taste, If you'd like to learn more and have a look in person, just pop into one of our stores and our expert staff will do their utmost to help you quit the ciggies for good! ================================ #photooftheday #photography #vapenation #vapeporn #vape #eliquid #eliquids #vaping #vapers #vapelyfe #vapefam #vapor #vapelife #ejuice #vaper #vapecommunity #vapedaily #instavape #ejuices #vapeon #vapestagram #vapepics #vapetricks #vapelove #cloudchaser #vapeshop #vapejuice #subohm | PRO | PRO |
| Heated tobacco products (HTPs) like IQOS and Eclipse, sometimes marketed as "heat-not-burn" technology, represent a diverse class of products that heat the tobacco leaf to produce an inhaled aerosol. They are different from e-cigarettes, which heat a liquid that can contain nicotine derived from tobacco. | NEU | NEU |
| Tobacco giant grapples with rules for marketing vaping product The authorities hold firm to the view that a hybrid tobacco product is still tobacco – and harmful – even if less harmful than smoking traditional cigarettes A store without a name -- and there are others. The no-name IQOS store sells vaping items but the catch here is that the vaping is done with tobacco rather than a liquid with (often) nicotine in it that is heated and turns to steam (rather than burns). Confused about the different products and tobacco companies' push for change? I tried to help walk readers through this confusing maze. Read my story. | ANTI | NEU |
| It is important that we continue to modernize proven tobacco prevention and control strategies to include newer products entering the market such as Heated Tobacco Products. | PRO | NEU |
| Making bioeconomy circular: How far can circular economy principles be applied to the bioeconomy? The European Commission has adopted an ambitious new Circular Economy Package to help European businesses and consumers make the | NEU | IR |
| transition to a stronger and more circular economy where resources are… Electronic cigarettes and heated tobacco products are a rapidly evolving category. Industry claims there is a need to regulate but in a proportionate manner. | | |
| **Twitter Messages** | **Human** | **GPT-4T** |
| 109 (2-24): New heated #tobacco device causes same damage to #lung cells as #e_cigs and #smoking | ANTI | ANTI |
| IQOS is billed as the most successful smoke-free product, with more than 12 million users around the world. Around 9 million IQOS users have completely stopped smoking. | PRO | PRO |
| 'Heat-not-burn' cigarettes on their way to U.S. market | NEU | NEU |
| "Heat-not-burn products, unlike their e-cigarette counterparts, do contain tobacco."<br>#cigarette 0<br>#cancer 1<br>So #vape https://t.co/BPZdf4Ye0U | ANTI | NEU |
| Too bad that they have no Iqos in ur shitty country | PRO | NEU |
| IQOS devices use a patented heat-control technology that precisely heats tobacco-filled sticks wrapped in paper, without the burning, to release a water-based aerosol – not smoke. | NEU | PRO |